\pdfoutput=1

\documentclass[11pt]{article}

\usepackage[final]{emnlp2021}

\usepackage{graphicx}
\usepackage{times}
\usepackage{latexsym}
\usepackage{multirow}
\usepackage{array}
\usepackage{algorithm2e}
\newcolumntype{P}[1]{>{\centering\arraybackslash}p{#1}}
\usepackage{pgfplots}

\usepackage[T1]{fontenc}

\usepackage[utf8]{inputenc}

\usepackage{microtype}

\usepackage{graphicx}
\usepackage{subcaption}
\usepackage{makecell}
\usepackage{multirow}
\usepackage{todonotes}
\usepackage{amsmath}
\usepackage{enumitem}
\usepackage{hyperref}
\usepackage{pgfplots}
\usepackage{caption}
\usepackage{amssymb}
\usepackage{algpseudocode}
\usepackage{tabularx,ragged2e}
\usepackage{lipsum}

\newcommand{\V}{\mathrm{V}}
\newcommand{\T}{\mathrm{T}}
\newcommand{\K}{\mathrm{K}}
\newcommand{\p}{\mathbf{p}}
\newcommand{\q}{\mathbf{q}}
\newcommand{\qprime}{\mathbf{q'}}
\newcommand{\sumnl}{\sum\nolimits}

\title{The Implicit Length Bias of Label Smoothing on Beam Search Decoding}

\author{Bowen Liang, Pidong Wang, Yuan Cao \\
        Google Research \\ 
        \texttt{\{bowenl,pidong,yuancao\}@google.com}}

\begin{document}
\maketitle
\begin{abstract}

Label smoothing is ubiquitously applied in Neural Machine Translation (NMT) training. While label smoothing offers a desired regularization effect during model training, in this paper we demonstrate that it nevertheless introduces length biases in the beam search decoding procedure. Our analysis shows that label smoothing implicitly applies a length penalty term to output sequence, causing a bias towards shorter translations. We also show that for a model fully optimized with label smoothing, translation length is implicitly upper bounded by a fixed constant independent of input. We verify our theory by applying a simple rectification function at inference time to restore the unbiased distributions from the label-smoothed model predictions. This rectification method led to consistent quality improvements on WMT English-German, English-French, English-Czech and English-Chinese tasks, up to +0.3 BLEU at beam size 4 and +2.8 BLEU at beam size 200.
\end{abstract}

\section{Introduction}

Label smoothing (LS) \cite{szegedy2016rethinking} is a simple yet effective approach that has been widely adopted in NMT training \cite{transformer, chen-etal-2018-best}. Acknowledging the ubiquitous uncertainty in label assignment, LS discounts the probability of ground-truth tokens in training data, and distributes the removed probability mass to remaining tokens in the vocabulary. Such softened target distributions alleviate the problem of overfitting the model to the observed (potentially noisy) labels during training, effectively improving its generalization ability.

The study of LS has been mostly focusing on how it shapes the model behavior during \emph{training} \cite{pereyra2017regularizing, muller2019when,  lukasik-etal-2020-semantic}, but one important yet overlooked aspect is how it affects the \emph{inference} procedure, especially for beam search, the most commonly used decoding algorithm for NMT. In this paper, we investigate this problem and elucidate that the application of LS skews the prediction distribution, introducing noise in sequence scoring, and biasing beam search towards shorter translations. 

To verify our analysis, we apply a simple debiasing operation to offset the bias during inference by rectifying distributions proposed by label-smoothed models. We conduct systematic experiments on WMT EnDe, EnFr, EnCs and EnZh tasks, and the empirical results resonate with our theoretical analysis: 

1) Beam search without debiasing indeed degrades translation quality, and this problem becomes more severe as the beam size increases.

2) LS-trained models are encouraged to generate shorter translations. This observation is aligned with \newcite{stahlberg-byrne-2019-nmt}, and our analysis may partially explain why short translation is a frequently encountered problem; 

3) Noise introduced by LS alone does not fully explain the length bias baked into the model. As we increase the strength of debiasing, we observed further quality gains even if it shifts away from the theoretically optimal value, which indicates stronger debiasing beyond offsetting LS noise is beneficial.

\section{Background}

\subsection{Label Smoothing}

LS is a commonly used technique that improves the generalization ability of NMT models. At training time, the model makes a prediction over the target-side vocabulary at each time step $t$, conditioned on source inputs $\mathbf{x}$ and prefix strings $y_{1\ldots{t-1}}$: 

\begin{align}
p^t_i \triangleq p(y^t_i|\mathbf{x}, y_{1...t-1}), i\in [1, 2, ... \V] \notag
\end{align}
where $\V$ is the size of vocabulary. The loss is usually the cross-entropy between $\mathbf{p}^t=[p^t_1,\ldots,p^t_V]$ and the ground-truth distribution $\mathbf{q}^t=[q^t_1,\ldots,q^t_V]$:
\small
\begin{equation}
L_t = - \sumnl_{i=1}^{\V} q^t_i \log (p^t_i) \notag
\end{equation}
\normalsize

To prevent overfitting on the hard 0-1 label distribution, LS softens the distribution via interpolation between $\mathbf{q}$ and a uniform distribution:

\small
\begin{equation} 
\qprime = (1 - \alpha) \q + \alpha / \V \label{eq:q}
\end{equation}
\normalsize
where $\alpha$ is a hyperparameter set to a small value. In NMT training, $\alpha$ is commonly set to 0.1.

\subsection{Beam Search}
Beam search is one of the most widely used algorithms for sequential model decoding. Beam search maintains the top $\K$ partial sequences of highest scores assigned by the model at every decoding step. At step $t$, the model uses the $\K$ partial sequences from step $t-1$ as prefixes to generate the probability distribution for the next token, obtaining $\K \cdot \V$ new extensions. The beam size $\K$ controls the width of search space: $\K=1$ is equivalent to greedy decoding, whereas $\K=\infty$ is equivalent to exact search.

Interestingly, it has been long observed that larger beam sizes lead to shorter translations and lower translation quality \cite{koehn-knowles-2017-six}. \newcite{stahlberg-byrne-2019-nmt} showed that with exact search, more than half of the outputs turn out to be empty translations, especially for long inputs. In the following section, we show that LS is partially responsible for this phenomenon.

\section{Bias and Debiasing in Beam Search}
\subsection{Label Smoothing Induces Bias in Beam Search}
The optimal distribution $\hat{\p}$ that minimizes the loss when LS is enabled is equivalent to the smoothened target distribution in Eq. \ref{eq:q} \cite{gao-etal-2020-towards}:

\small
\begin{equation} \label{eq:1}
\hat{\p} = \qprime = (1 - \alpha) \q + \alpha / \V
\end{equation}
\normalsize

It is worth noting that while the label distribution $\q$ in Eq. \ref{eq:q} is the 0-1 hard label for a specific appearance of a token in training data, the $\q$ here should be understood as the linguistic ground truth that minimizes the global loss of the whole dataset. This will be a soft distribution when intrinsic uncertainty of the NMT task \cite{ott:uncertainty:2018} is considered.

During beam search, since only top-$\K$ candidates are being considered at each step, their probabilities dominate the procedure and the $\alpha / \V$ term becomes insignificant. We can therefore approximate the log probability of the target sequence as:

\small
\begin{align*}
\sum_{t=1}^\T \log \hat{p} (y_t) 
    & = \sum_{t=1}^\T \log ((1 - \alpha) q(y_t) + \alpha / \V ) \\
    & \approx \sum_{t=1}^\T \log ((1 - \alpha) q(y_t)) \\
    & = \sum_{t=1}^\T \log q(y_t) + \T \log (1 - \alpha) \notag
\end{align*}
\normalsize

Compared to $\sum_{t} \log q(y_t)$ which is the true likelihood of the sequence, the model prediction $\sum_{t} \log \hat{p} (y_t)$ effectively applies a penalty term $\log (1 - \alpha)$ to every target token, artificially reducing the likelihood of longer sequences hence causing a bias towards shorter outputs. When $\alpha =0.1$, the per-token penalty equals $\log 0.9 \approx -0.105$.

\subsection{Upper Bound of Output Length} \label{sec:upper}
Another perspective to look at the bias problem is given by the examination of upper and lower bounds of sequence likelihoods. The probability of any token assigned by a perfectly optimized model is $\hat{p_i} = (1 - \alpha) q_i + \alpha / \V \in [\alpha / \V, 1 - \alpha + \alpha / \V]$. As a result, we have the following bounds for empty translations (predicting EOS at the first step) and translations of length $\T$:

\small
\begin{equation}
\begin{aligned}
& \hat{p}(\mathrm{EOS}|\mathbf{x}) \geq \alpha / \V \\
& \hat{p}(y_{1...\T}|\mathbf{x}) = \prod_{t=1}^{\T} \hat{p}(y_t) \leq (1 - \alpha + \alpha / \V)^{\T} \notag
\end{aligned}
\end{equation}
\normalsize

As an example, for $\alpha=0.1$ and $\V=32000$, since $0.1/32000 > (1 - 0.1 + 0.1/32000)^{121}$, any translations longer than 121 tokens are bound to have lower scores than empty translations. The general argument here is that LS biases search results by implicitly upper-bounding the output length to $\T_{max} = \log(\alpha / \V) / \log (1 - \alpha + \alpha / \V) .$

\subsection{Debiasing from Label Smoothing}
Having identified the bias induced by LS, we now introduce an operation to recover the unbiased distribution $\q$ from the learned $\hat{\p}$ at inference time:

\small
\begin{equation} \label{eq:3}
  \p^{db} = \q = ( \hat{\p} - \alpha / \V ) / (1 - \alpha)
\end{equation}
\normalsize

where $\p^{db}$ stands for ``debiased'' model prediction. This equation is obtained by rearranging Eq. \ref{eq:1}. According to the analysis from Sec.\ref{sec:upper}, $\hat{p_i}$ is in $[\alpha / \V, 1 - \alpha + \alpha / \V]$, hence the output $p_i^{db}$ will be in $[0.0, 1.0]$ and is guaranteed to sum to 1.

In practice however, due to limited model expressiveness and imperfect optimization, outlier values of $\hat{p_i} < \alpha / \V$ can be observed, resulting in negative $p_i^{db}$. To avoid this problem we apply a ReLU operation to $p_i^{db}$ and normalize their sum to 1: 

\small
\begin{equation}
\begin{aligned}
\label{eq:8}
p_i^{db} &= \frac{ \mathrm{ReLU}(\hat{p_i} - \delta) / (1 - \alpha) }{\sum_{j=1}^{\V} \mathrm{ReLU}(\hat{p_j} - \delta) / (1 - \alpha)} \\
&= \frac{ \mathrm{ReLU}(\hat{p_i} - \delta) }{\sum_{j=1}^{\V} \mathrm{ReLU}(\hat{p_j} - \delta) }
\end{aligned}
\end{equation}
\normalsize

where $\delta$ is debiasing parameter. When $\delta = \alpha / \V$ and $\hat{p_i} \geq \delta$, this definition is equivalent with Eq. \ref{eq:3}. Note that the $/(1-\alpha)$ term has been dropped since it has no effect under normalization.

Although the theoretically optimal value for $\delta$ is $\alpha / \V$, in experiments we show that increasing $\delta$ above $\alpha / \V$ can yield even larger empirical improvements. This indicates that the proposed technique has benefits in offsetting noises from other sources beyond label smoothing.

\section{Experiments}

\subsection{Setups}

We conduct experiments on WMT19 EnDe, EnCs, EnZh, and WMT15 EnFr. For each task, we train a Transformer-Base model \cite{transformer} with LS $\alpha=0.1$, and decode with different beam sizes and $\delta$ values in Eq. \ref{eq:8}. In addition to the $\delta=0$ baseline (no debiasing), we also train a ``No-LS'' baseline without label smoothing.

All models are trained with minibatches of 256k tokens for 600k steps. The training data is preprocessed using SentencePiece model \cite{kudo-richardson-2018-sentencepiece} with $\V = 32000$. We conduct experiments with the open-source Lingvo framework \cite{shen2019lingvo}, and use SacreBLEU \cite{sacrebleu} as quality measure. We did not employ length normalization during beam search. While length normalization can empirically alleviate quality loss at large beam sizes \cite{murray-chiang-2018-correcting}, it does not provide an explanation to the modelling biases observed in \newcite{stahlberg-byrne-2019-nmt}, and that is what we are trying to address in this paper.

\subsection{Results}

\begin{figure}
\centering

\begin{tikzpicture}[scale=0.7]
\begin{axis}[
    xlabel={Debiasing Parameter $\delta$},
    xmin=1, xmax=5,
    xtick={1,2,3,4,5},
    xticklabels={0,0.1/V,1/V,10/V,100/V},
    width=9cm, height=6cm,
    ymin=37.0, ymax=40.5,
    ytick={37.0,37.5,38.0,38.5,39.0,39.5,40.0,40.5},
    yticklabel style={%
                 /pgf/number format/.cd,
                     fixed,
                     fixed zerofill,
                     precision=1,
                     },
    legend pos=outer north east,
    ymajorgrids=true,
    grid style=dashed,
    ]
\addplot[color=orange,
    mark=+,
    ] table [] {
1	39.28
2.00	39.28
2.48	39.28
2.70	39.28
3.00	39.28
3.30	39.28
3.60	39.28
4.00	39.28
4.20	39.28
4.51	39.28
4.81	39.28
5.00	39.28
};
\addlegendentry{K=1}
\addplot[color=blue,
    mark=square,
    ] table [] {
1	40.02
2.00	40.05
2.48	40.18
2.70	40.19
3.00	40.2
3.30	40.23
3.60	40.28
4.00	40.25
4.20	40.27
4.51	40.25
4.81	40.26
5.00	40.25
};
\addlegendentry{K=4}
\addplot[color=brown,
    mark=asterisk,
    ] table [] {
1	39.89
2.00	39.97
2.48	40.07
2.70	40.07
3.00	40.03
3.30	40.07
3.60	40.04
4.00	40.07
4.20	40.04
4.51	40.05
4.81	39.96
5.00	40.01
};
\addlegendentry{K=8}
\addplot[color=violet,
    mark=diamond]
    table [] {
1	39.14
2.00	39.42
2.48	39.44
2.70	39.45
3.00	39.48
3.30	39.49
3.60	39.58
4.00	39.69
4.20	39.64
4.51	39.58
4.81	39.64
5.00	39.7
};
\addlegendentry{K=25}
\addplot[color=black,
    mark=o]
    table [] {
1	38.35
2.00	38.77
2.48	38.92
2.70	38.94
3.00	38.94
3.30	38.97
3.60	39.06
4.00	39.11
4.20	39.1
4.51	39.16
4.81	39.29
5.00	39.36
};
\addlegendentry{K=100}
\addplot[color=red,
    mark=triangle]
    table [] {
1	37.24
2.00	37.98
2.48	38.22
2.70	38.52
3.00	38.5
3.30	38.64
3.60	38.79
4.00	38.98
4.20	38.97
4.51	39.05
4.81	39.25
5.00	39.3
};
\addlegendentry{K=200}
\end{axis}
\end{tikzpicture}
\caption{EnDe BLEU for various $\delta$ and beam size $\K$}
\label{fig:bleu-per-db}
\end{figure}
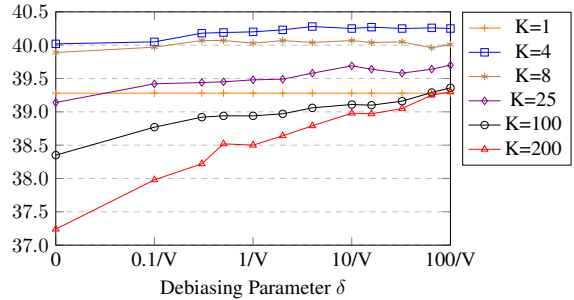

Figure \ref{fig:bleu-per-db} shows the EnDe BLEU scores for varied debiasing parameters. At beam size 4, setting $\delta \geq 0.5/\V$ leads to $+0.2$ improvements over the $\delta=0$ baseline. At beam size 200, increasing $\delta$ consistently improves BLEU, with $+2.1$ gain at $\delta = 100/\V$. It is not surprising that debiasing has a larger impact for large beam sizes, as it does not change the relative ranking of candidate tokens within each step, but only affects the ranking of (partial or complete) sequences. In the extreme case of greedy decoding, debiasing has no effect. 

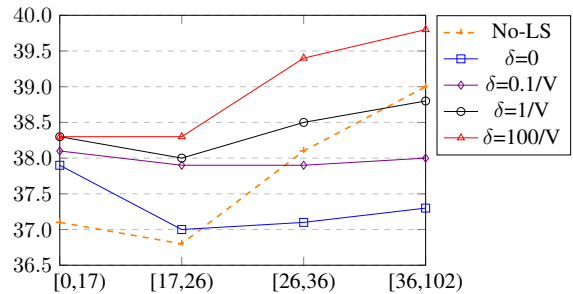
\begin{figure}
\centering
\begin{tikzpicture}[scale=0.75]
\begin{axis}[
    xmin=1, xmax=4,
    width=8cm, height=6cm,
    xtick={1,2,3,4},
    xticklabels={{\hspace{7mm}[0,17)},{[17,26)},{[26,36)},{[36,102)}},
    ymin=36.5, ymax=40.0,
    ytick={36.5,37.0,37.5,38.0,38.5,39.0,39.5,40.0},
    yticklabel style={%
                 /pgf/number format/.cd,
                     fixed,
                     fixed zerofill,
                     precision=1,
                     },
    legend pos=outer north east,
    ymajorgrids=true,
    grid style=dashed,
    ]
\addplot[color=orange,
    mark=+, dashed, style=thick]
    table [] {
x    y       label   alignment
1    37.1    37.1    +145
2    36.8    36.8    +90
3    38.1    38.1    +180
4    39.0    39.0    -30
};
\addlegendentry{No-LS}
\addplot[color=blue,
    mark=square]
    table [] {
x    y       label   alignment
1    37.9    37.9    +150
2    37.0    37.0    -90
3    37.1    37.1    -90
4    37.3    37.3    -30
};
\addlegendentry{$\delta$=0}
\addplot[color=violet,
    mark=diamond]
    table [] {
x    y       label   alignment
1    38.1    38.1    +160
2    37.9    37.9    +100
3    37.9    37.9    +90
4    38.0    38.0    +30
};
\addlegendentry{$\delta$=0.1/V}
\addplot[color=black,
    mark=o]
    table [] {
x    y       label   alignment
1    38.3    38.3    +170
2    38.0    38.0    +180
3    38.5    38.5    -90
4    38.8    38.8    +35
};
\addlegendentry{$\delta$=1/V}
\addplot[color=red,
    mark=triangle]
    table [] {
x    y       label   alignment
1    38.3    38.3    -150
2    38.3    38.3    +180
3    39.4    39.4    -90
4    39.8    39.8    +30
};
\addlegendentry{$\delta$=100/V}
\end{axis}
\end{tikzpicture}
\caption{EnDe BLEU per source length. Buckets are chosen to evenly split the test set. Beam size = 200. }
\label{fig:bleu-per-length-bucket}
\end{figure}

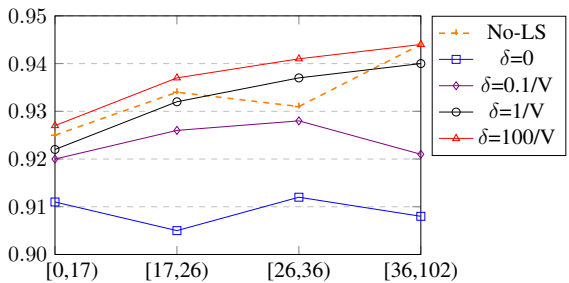
\begin{figure}
\centering
\begin{tikzpicture}[scale=0.75]
\begin{axis}[
    xmin=1, xmax=4,
    width=8cm, height=5.8cm,
    xtick={1,2,3,4},
    xticklabels={{\hspace{6mm}[0,17)},{[17,26)},{[26,36)},{[36,102)}},
    ymin=0.90, ymax=0.95,
    ytick={0.90,0.91,0.92,0.93,0.94,0.95},
    yticklabel style={%
                 /pgf/number format/.cd,
                     fixed,
                     fixed zerofill,
                     precision=2,
                     },
    legend pos=outer north east,
    ymajorgrids=true,
    grid style=dashed,
    ]
\addplot[color=orange,
    mark=+, dashed, style=thick]
    table [] {
x    y       label   alignment
1    0.925    .925    -170
2    0.934    .934    -160
3    0.931    .931    +180
4    0.944    .944    -30
};
\addlegendentry{No-LS}
\addplot[color=blue,
    mark=square]
    table [] {
x    y       label   alignment
1    0.911    .911    +150
2    0.905    .905    -90
3    0.912    .912    -90
4    0.908    .908    -30
};
\addlegendentry{$\delta$=0}
\addplot[color=violet,
    mark=diamond]
    table [] {
x    y       label   alignment
1    0.920    .920    +160
2    0.926    .926    +100
3    0.928    .928    +90
4    0.921    .921    +30
};
\addlegendentry{$\delta$=0.1/V}
\addplot[color=black,
    mark=o]
    table [] {
x    y       label   alignment
1    0.922    .922    +180
2    0.932    .932    +90
3    0.937    .937    -140
4    0.940    .940    +35
};
\addlegendentry{$\delta$=1/V}
\addplot[color=red,
    mark=triangle]
    table [] {
x    y       label   alignment
1    0.927    .927    -140
2    0.937    .937    -90
3    0.941    .941    -90
4    0.944    .944    -30
};
\addlegendentry{$\delta$=100/V}
\end{axis}
\end{tikzpicture}
\caption{EnDe translation/reference length ratio per source length bucket, beam size = 200}
\label{fig:len-ratio-per-length-bucket}
\end{figure}

Figure \ref{fig:bleu-per-length-bucket} shows the BLEU bucketed by source lengths, decoded at beam size 200. No-LS underperforms label-smoothed model on shorter queries, but outperforms on long ones. This is in accordance with our theory, as the effect of the $\T \log (1 - \alpha)$ bias term grows proportionally to sequence length. Figure \ref{fig:len-ratio-per-length-bucket} examines the length ratio of each system. No-LS produces significantly longer translations than the LS model, which is also aligned with our theoretical analysis. By setting the debiasing parameter $\delta>0$, BLEU and length ratio are improved across all length buckets.

\begin{table*}[bt!]
\centering
\begin{tabular}{ P{1.5cm} P{0.8cm} P{0.8cm} P{0.8cm} P{0.8cm} P{0.8cm} P{0.8cm} P{0.8cm} P{0.8cm} P{0.8cm} }
\hline
 & \multirow{2}{*}{Beam} & \multirow{2}{*}{No-LS} & \multicolumn{6}{c}{LS with debiasing parameter $\delta$}  \\ 
 &  &  & 0 & $0.1/\V$ & $0.5/\V$ & $1/\V$ & $10/\V$ & $100/\V$  \\ 
\hline
 EnCs & 4 & 24.2 & 24.5 & \bf{24.6} & 24.5 & 24.5 & 24.5 & 24.3  \\ 
 EnCs & 200 & 23.4 & 23.4 & 23.6 & 23.6 & 23.7 & 24.0 & \bf{24.1}  \\ 
 \hline
 EnDe & 4 & \bf{40.3} & 40.0 & 40.0 & 40.2 & 40.2 & \bf{40.3} & 40.2 \\ 
 EnDe & 200 & 38.2 & 37.2 & 38.0 & 38.5 & 38.5 & 39.0 & \bf{39.3} \\ 
\hline
 EnFr & 4   & 38.9 & 39.1 & 39.1 & \bf{39.2} & \bf{39.2} & \bf{39.2} & \bf{39.2} \\ 
 EnFr & 200 & 35.5 & 35.6 & 36.5 & \bf{36.7} & 36.6 & 36.1 & 36.6 \\ 
\hline
 EnZh & 4   & 28.6 & \bf{29.5} & \bf{29.5} & \bf{29.5} & \bf{29.5} & \bf{29.5} & 29.4 \\ 
 EnZh & 200 & 24.8 & 23.7 & 24.0 & 24.4 & 24.6 & 25.7 & \bf{26.5} \\ 
\hline
\end{tabular}
\vspace*{-2mm}
\caption{BLEU for varied beam sizes and languages}
\label{tab:bleu_len_ratios}
\end{table*}

\begin{table*}[bt!]
\centering
\begin{tabular}{ P{0.7cm} P{4cm} P{1cm} P{1cm} P{1cm} P{1cm} P{1cm} P{1cm} P{1cm} P{1cm}}
\hline
 & \multirow{2}{*} & \multirow{2}{*}{No-LS} & \multicolumn{6}{c}{LS with debiasing parameter $\delta$}  \\ 
 &  &  & 0 & $0.1/\V$ & $0.5/\V$ & $1/\V$ & $10/\V$ & $100/\V$  \\ 
\hline
EnDe & Sum prob of $S$ & 0.097 & 0.031 & 0.067 & 0.090 & 0.096 & \bf{0.121} & 0.179 \\
EnDe & Ratio of reference in $S$ & 0.136 & 0.136 & 0.136 & 0.136 & 0.136 & \bf{0.134} & 0.133 \\
\hline
EnFr & Sum prob of $S$ & 0.098 & 0.025 & 0.067 & 0.088 & \bf{0.094} & 0.118 & 0.181 \\
EnFr & Ratio of reference in $S$ & 0.094 & 0.095 & 0.094 & 0.095 & \bf{0.095} & 0.093 & 0.094 \\
\hline
\end{tabular}
\vspace*{-1mm}
\caption{Set-level calibration for EnDe and EnFr. $S$ represents the set of top 200 beam search results. Bold font marks the system with best calibration accuracy.}
\label{tab:calibration}
\end{table*}

Table \ref{tab:bleu_len_ratios} shows BLEU for 4 language pairs at beam size 4 and 200. With beam size 4, $\delta=1/\V$ provides near-peak performance for all language pairs. This is intuitively reasonable, since $1/\V$ corresponds to the probability of blindly picking a choice out of the vocab, and $\delta = 1/\V$ effectively forces the model to discard ``blind'' choices. With beam size 200, large $\delta$ values consistently improve quality, with up to +2.8 BLEU on EnZh.

\subsubsection{Measuring Confidence with Calibration}
By enabling LS during training, one hopes that it can prevent the model from being over-confident. With the introduction of debiasing which reverts the effect of LS, a natural question to ask is whether debiasing will result in over-confident model predictions. We answer this question via Set-Level Calibration \cite{ott:uncertainty:2018} analysis. More specifically, for each query we sum the prediction probability of the top 200 beam search outputs, denoted by $S$, and compare it with the actual frequency that the reference is contained in $S$. These two numbers should match closely if a model is well calibrated.

Tabel \ref{tab:calibration} presents the set-level calibration result on EnDe and EnFr. The baseline $\delta = 0$ system is severely under-confident, with the predicted probability being < 1/3 of observed frequency. The No-LS system, in comparison, has much better calibration accuracy. This contradicts the results from \newcite{muller2019when}, which observed the LS system to be better calibrated on token level. Our debiasing method achieves high calibration accuracy with $\delta=10/\V$ for EnDe and $\delta=1/\V$ for EnFr, and becomes over-confident at $\delta=100/\V$.

\section{Related Work}

Previous studies focused on understanding and improving the effects of LS in training. \newcite{muller2019when} analyzed its effect on internal representations and model calibration. \newcite{pmlr-v119-lukasik20a, chen2020investigation} investigated its connection with label noise. \newcite{pereyra2017regularizing, lukasik-etal-2020-semantic, wang-etal-2020-inference, gao-etal-2020-towards, meister-etal-2020-generalized, peters-martins-2021-smoothing} proposed more sophisticated smoothing distribution to replace the uniform distribution. While previous research primarily focused on model training, our work focuses on performance improvement during inference.

The problem of ``Beam search degradation'' has received much attention since \newcite{koehn-knowles-2017-six}. A few methods have been proposed to mitigate this issue, including length normalization \cite{jean-etal-2015-montreal, wu2016googles}, word reward \cite{he-2016-snmt-features, yang-etal-2018-breaking, murray-chiang-2018-correcting}, length prediction \cite{yang-etal-2020-predicting} and improved search objective \cite{pmlr-v97-cohen19a, meister-etal-2020-beam}. \newcite{ott:uncertainty:2018} studied the extrinsic noise in training data, and concluded that the copying pattern in training data contributed to beam search degradation. Nevertheless, \newcite{stahlberg-byrne-2019-nmt} pointed out that it remains an open question why NMT models assign high probabilities to empty translations. More recently, \newcite{shi-2020-nmt-empty-outputs} and \newcite{kulikov-eremeev-oversmooth} also observed the connection between empty translation, label smoothing, and oversmoothing. However, to the best of our knowledge, our work is the first to provide a quantitative analysis on the biasing effect of Label Smoothing and propose a solution to counteract the bias.

\section{Conclusion}
We demonstrate that Label Smoothing implicitly introduces biases in beam search and encourages shorter translations. Our analysis led to a disturbing conclusion that, under infinite beam size, LS-trained models are incapable of producing sequences longer than 121 tokens (assuming 32k vocab and $\alpha = 0.1$), regardless of input length. Our experiments show that with large beam sizes, translations produced by LS models indeed suffer from shorter length and worse quality than No-LS models, which confirms our theoretical analysis.

We propose a simple rectification method to offset the bias introduced by LS at inference time, which has been shown to have effectively improved beam search quality, especially under large beam sizes. Furthermore, our methods indicate that the target distributions for training can be different from the optimal distributions for inference, which is an interesting topic for future research.

\section*{Acknowledgements}

We thank Weikang Zhou, Wolfgang Macherey, and Macduff Hughes for their insightful discussions; and thank the anonymous reviewers for their kind advices.

\bibliography{anthology,custom}
\bibliographystyle{acl_natbib}

\appendix

\end{document}